\documentclass{article}

\usepackage{PRIMEarxiv}

\usepackage[utf8]{inputenc} 
\usepackage[T1]{fontenc}    
\usepackage{hyperref}       
\usepackage{url}            
\usepackage{booktabs}       
\usepackage{amsfonts}       
\usepackage{nicefrac}       
\usepackage{microtype}      
\usepackage{lipsum}
\usepackage{graphicx}
\usepackage{float}
\usepackage{amsmath}
\usepackage{listings}
\usepackage{geometry}
\usepackage{xcolor}
\usepackage{algorithm}
\usepackage{algpseudocode}
\usepackage{fontawesome5}
\graphicspath{{media/}}     

\definecolor{codegray}{rgb}{0.95,0.95,0.95}

\title{LoongFlow: Directed Evolutionary Search via a Cognitive Plan-Execute-Summarize Paradigm}

\author{
  \textbf{Chunhui Wan}\thanks{Equal contributions. E-mail: \texttt{\{daixunan, wangzhuo12\}@baidu.com}} \quad
  \textbf{Xunan Dai}\textsuperscript{*} \quad
  \textbf{Zhuo Wang}\textsuperscript{*} \quad
  \textbf{Minglei Li}\textsuperscript{*} \quad
  \textbf{Yanpeng Wang}\textsuperscript{*} \quad
  \AND
  \textbf{Yinan Mao} \quad
  \textbf{Yu Lan} \quad
  \textbf{Zhiwen Xiao} \quad
  \AND
  Baidu Inc. 
}

\begin{document}
\maketitle

\begin{abstract}
The transition from static Large Language Models (LLMs) to self-improving agents is hindered by the lack of structured reasoning in traditional evolutionary approaches. Existing methods often struggle with premature convergence and inefficient exploration in high-dimensional code spaces. To address these challenges, we introduce \textbf{LoongFlow}, a self-evolving agent framework that achieves state-of-the-art solution quality with significantly reduced computational costs. Unlike "blind" mutation operators, LoongFlow integrates LLMs into a cognitive \textbf{"Plan-Execute-Summarize" (PES)} paradigm, effectively mapping the evolutionary search to a reasoning-heavy process. To sustain long-term architectural coherence, we incorporate a hybrid evolutionary memory system. By synergizing Multi-Island models with MAP-Elites and adaptive Boltzmann selection, this system theoretically balances the exploration-exploitation trade-off, maintaining diverse behavioral niches to prevent optimization stagnation. We instantiate LoongFlow with a \textbf{General Agent} for algorithmic discovery and an \textbf{ML Agent} for pipeline optimization. Extensive evaluations on the AlphaEvolve benchmark and Kaggle competitions demonstrate that LoongFlow outperforms leading baselines (e.g., OpenEvolve, ShinkaEvolve) by up to 60\% in evolutionary efficiency while discovering superior solutions. LoongFlow marks a substantial step forward in autonomous scientific discovery, enabling the generation of expert-level solutions with reduced computational overhead.

\vspace{0.5cm}
\centering
\faGithub\ \textbf{Code:} \url{https://github.com/baidu-baige/LoongFlow}
\end{abstract}

\section{Introduction}

The progression from static prompting—where humans manually engineer instructions—to autonomous, self-evolving agents marks a fundamental shift in artificial intelligence. While static approaches rely on fixed inference patterns, self-evolving agents utilize Large Language Models (LLMs) as mutation operators to iteratively modify their own code or parameters. Pioneering works have validated this paradigm in specific domains: FunSearch~\cite{romera2024mathematical} utilizes LLMs to discover novel mathematical constructions, Eureka~\cite{ma2024eureka} optimizes reward functions via evolutionary search, and AlphaEvolve~\cite{novikov2025alphaevolve} automates the discovery of heuristic algorithms. that LLMs, building on foundational code-generation capabilities~\cite{chen2021codex, li2022alphacode}, can discover novel mathematical algorithms and reward functions that surpass human baselines. This "Darwinian shift" has established automated scientific discovery as a vibrant research frontier.

However, as the complexity of tasks increases, current frameworks face severe cognitive and architectural limitations. Leading open-source baselines, such as OpenEvolve and ShinkaEvolve~\cite{sakana2025shinkaevolve}, effectively treat the LLM as a stochastic black box. OpenEvolve relies on high-volume random mutations, leading to a "random walk" behavior that is computationally prohibitive. ShinkaEvolve improves efficiency via novelty search but operates purely at the execution level, lacking a mechanism to analyze \textit{why} a mutation failed. Consequently, these methods hit a "cognitive ceiling," struggling to maintain structural coherence over long evolutionary horizons. Specifically, they encounter three critical bottlenecks:

\begin{itemize}
    \item \textbf{Inefficient Exploration (The Cost Bottleneck)}: Existing agents lack a strategic planning layer. They engage in brute-force sampling in high-dimensional code spaces, resulting in excessive token consumption and unstable convergence rates.
    
    \item \textbf{Diversity Collapse (The Convergence Bottleneck)}: Without explicit diversity management, population-based agents tend to converge prematurely to local optima. Traditional "Top-K" sampling fails to preserve diverse but potentially high-reward "stepping stone" solutions.
    
    \item \textbf{Absence of Reflexive Memory (The Feedback Bottleneck)}: Unlike deep learning, where backpropagation provides a precise gradient for improvement, most existing evolutionary agent frameworks lack a structured reflection mechanism~\cite{shinn2023reflexion}. They function as "memory-less" searchers, repeating similar errors across generations rather than accumulating "evolutionary wisdom" through structured summarization.
\end{itemize}

\begin{figure}[htbp]
  \centering
  \includegraphics[width=0.9\linewidth]{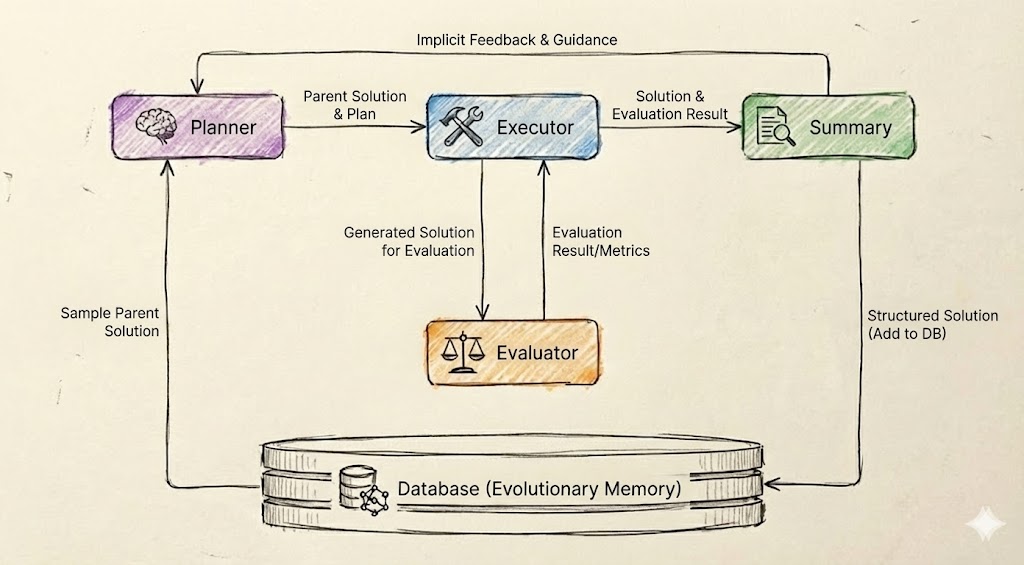}
  \caption{Overview of LoongFlow.}
  \label{fig:fig1}
\end{figure}

To overcome these barriers, we introduce \textbf{LoongFlow}, a framework designed to bridge the gap between reasoning agents and evolutionary computation. LoongFlow distinguishes itself through two core architectural innovations. First, we propose the \textbf{"Plan-Execute-Summarize" (PES)} paradigm. This cognitive loop transforms random mutation into a directed hypothesis-testing process (addressing Bottleneck 1 \& 3), as illustrated in the \textit{Agent Loop} of Figure~\ref{fig:fig1}. Second, to resolve the exploration-exploitation dilemma (Bottleneck 2), we design a \textbf{Hybrid Evolutionary Memory}. By fusing the spatial isolation of Island Models with the behavioral diversity of MAP-Elites~\cite{mouret2015illuminating} and entropy-regularized Boltzmann selection, LoongFlow dynamically maintains diverse behavioral niches. This ensures that the system can escape local optima and continuously discover novel solution architectures.

We demonstrate the versatility of LoongFlow by instantiating two domain-specific agents: a \textbf{General Agent} for algorithmic tasks and an \textbf{ML Agent} for machine learning pipelines. Experimental results on the AlphaEvolve benchmark and Kaggle competitions confirm that LoongFlow significantly surpasses OpenEvolve and ShinkaEvolve, breaking theoretical performance barriers with superior sample efficiency.

The primary contributions of this work are as follows:
\begin{itemize}
    \item \textbf{Structured Evolutionary Paradigm}: We propose the ``Plan-Execute-Summarize'' paradigm, which integrates expert-level planning and retrospective summarization to reduce generation randomness and establish a sustainable feedback loop.
    
    \item \textbf{Advanced Memory Architecture}: We design a domain-adaptive evolutionary memory that combines multi-island parallel evolution with MAP-Elites~\cite{mouret2015illuminating} and adaptive Boltzmann selection, effectively solving the premature convergence and ``catastrophic forgetting'' problems inherent in traditional LLM agents.
    
    \item \textbf{Superior Performance \& Efficiency}: We provide open-source, pre-built agents (GeneralAgent and MLAgent) that achieve state-of-the-art results on NP-hard mathematical problems and complex ML pipelines, surpassing existing frameworks in both stability and evolutionary speed.
\end{itemize}

\section{Related Work}

The development of LoongFlow is situated at the intersection of \textbf{LLM-based Evolutionary Optimization} and \textbf{Cognitive Agent Architectures}. In this section, we review the progression of these fields and identify the specific gap that LoongFlow addresses.

\subsection{LLM-Based Evolutionary Optimization}
The paradigm of utilizing LLMs for evolutionary optimization has shifted from simple solution generation to iterative refinement. Pioneering work such as FunSearch~\cite{romera2024mathematical} demonstrated that LLMs, when coupled with an evolutionary evaluator, could solve open problems in mathematics (e.g., the Cap Set problem) by searching for "functions" rather than parameters. Similarly, AlphaEvolve~\cite{novikov2025alphaevolve} orchestrates an autonomous pipeline of LLMs to improve an algorithm by making direct changes to the program, achieving human-level performance on robot manipulation tasks. While recent methods like OPRO~\cite{yang2023opro} and PromptBreeder~\cite{fernando2023promptbreeder} have explored using LLMs as optimizers, they often treat the model as a black-box operator to mutation. They typically treat the LLM as a stochastic operator—randomly mutating code without a high-level strategy—which leads to high token costs and inefficient exploration in complex search spaces.

\subsection{Evolutionary Agent Frameworks}
To engineeringly scale LLM-based evolution, several open-source frameworks have emerged. These serve as the primary baselines for our work:

\begin{itemize}
    \item \textbf{OpenEvolve}: As a standard implementation of the AlphaEvolve algorithm, OpenEvolve utilizes an "Island Model" to maintain population diversity. However, it treats code generation as a single-step translation task. The mutation process is largely reactive, where the agent fixes errors or makes random local changes without understanding the global algorithmic structure. This often leads to a "random walk" behavior in high-difficulty tasks.
    
    \item \textbf{ShinkaEvolve}: ShinkaEvolve~\cite{sakana2025shinkaevolve} improves sample efficiency by integrating "code-novelty rejection" to filter out redundant solutions before execution. While it reduces computational waste, it still operates primarily at the execution level. The framework lacks a structured reflection mechanism to analyze \textit{why} a specific architectural change failed, preventing the system from learning abstract principles over long evolutionary horizons.
\end{itemize}

\subsection{Cognitive Architectures and The Reasoning Gap}
While evolutionary methods excel at population-based search, they often lack the depth of semantic reasoning found in autonomous agents.

\textbf{Reasoning Agents}: Frameworks like ReAct~\cite{yao2023react} and Reflexion~\cite{shinn2023reflexion} have demonstrated that interleaving reasoning traces (Thought) with actions significantly improves problem-solving capabilities. These methods enable agents to perform multi-step planning and self-correction. Similarly, Voyager~\cite{wang2023voyager} utilizes an iterative curriculum to learn complex skills in embodied environments.

\textbf{The Gap}: A critical gap exists in merging these two paradigms. Standard reasoning agents (like AutoGPT~\cite{richards2023autogpt} or Voyager~\cite{wang2023voyager}) generally focus on single-instance problem solving rather than population-based evolutionary search. Conversely, traditional evolutionary algorithms (like MAP-Elites~\cite{mouret2015illuminating}) excel at maintaining diverse populations but lack the semantic reasoning capabilities of ReAct-style agents.

\textbf{LoongFlow} bridges this gap by introducing the \textbf{"Plan-Execute-Summarize"} paradigm. This paradigm allows LoongFlow to maintain the diversity benefits of MAP-Elites~\cite{mouret2015illuminating} while leveraging the reasoning depth of ReAct-style agents~\cite{yao2023react}, effectively moving the evolutionary process from "random mutation" to "directed evolution".

\section{Background}
In this section, we formally frame the open-ended evolutionary process as a sequential decision-making problem and establish the mathematical foundations of the LoongFlow framework. We model the self-evolution of agents as a Markov Decision Process (MDP~\cite{sutton2018reinforcement}) over a discrete code space, guided by a parameterized Large Language Model (LLM).

\subsection{Problem Formulation as MDP}
We define the problem as a tuple $\langle \mathcal{C}, \mathcal{A}, R, \pi_\theta \rangle$:
\begin{itemize}
    \item \textbf{State/Code Space ($\mathcal{C}$)}: Let $\mathcal{C}$ be the infinite, discrete space of all valid programs in a specific language (e.g., Python). A state $s_t \in \mathcal{C}$ represents the solution code at evolutionary generation $t$.
    \item \textbf{Action Space ($\mathcal{A}$)}: The action space consists of semantic modification operations (e.g., rewrite, debug, optimize) applied to the code.
    \item \textbf{Reward Function ($R$)}: $R: \mathcal{C} \to \mathbb{R}$ is a scalar fitness function (e.g., accuracy on test cases). The environment is characterized by a sparse reward signal, where valid solutions are rare.
    \item \textbf{Policy ($\pi_\theta$)}: The agent is an LLM parameterized by weights $\theta$. It acts as a stochastic policy $\pi_\theta(a|s)$, generating the next code state $s_{t+1}$ based on the current state $s_t$ and context.
\end{itemize}

Our objective is to find an optimal solution $s^*$ that maximizes the reward:
\begin{equation}
    s^* = \mathop{\arg\max}_{s \in \mathcal{C}} R(s)
\end{equation}

\subsection{LLM as a Composite Semantic Operator}
Unlike traditional Evolutionary Algorithms (EA) that use fixed, random mutation operators (denoted as $\mathcal{T}_{mut}$), LoongFlow utilizes the LLM as a learnable \textbf{Semantic Operator}.

We formalize the "Plan-Execute-Summarize" (PES) paradigm as a composite transition kernel decomposing the policy $\pi_\theta$ into three sub-steps. Let $\mathcal{M}_t$ be the evolutionary memory at generation $t$, and $\mathcal{I}$ be the set of system instructions (prompts). The transition from parent $s_t$ to offspring $s_{t+1}$ proceeds as follows:

\begin{enumerate}
    \item \textbf{Planning}: The Planner generates a natural language blueprint $b$ (an intermediate latent variable) conditioned on the parent code $s_t$ and retrieved insights from memory $\mathcal{M}_t$:
    \begin{equation}
        b \sim \pi_\theta(b \mid s_t, \mathcal{M}_t, \mathcal{I}_{plan})
    \end{equation}
    
    \item \textbf{Execution}: The Executor generates the executable offspring code $s'$ (a candidate for $s_{t+1}$) based on the blueprint $b$:
    \begin{equation}
        s' \sim \pi_\theta(s' \mid b, s_t, \mathcal{I}_{exec})
    \end{equation}
    
    \item \textbf{Summarization \& Update}: The Summarizer generates a reflection insight $z$ based on the execution feedback $r = R(s')$, and updates the memory:
    \begin{equation}
        z \sim \pi_\theta(z \mid s', r, b, \mathcal{I}_{sum})
    \end{equation}
    \begin{equation}
        \mathcal{M}_{t+1} \leftarrow \mathcal{M}_t \cup \{z\}
    \end{equation}
\end{enumerate}

\subsection{Feature Space and Archive Management}
To manage population diversity beyond raw fitness, we map the high-dimensional code space $\mathcal{C}$ to a lower-dimensional \textbf{Feature Space} $\mathcal{F} \subseteq \mathbb{R}^k$.

\paragraph{Feature Mapping.} Let $\Phi: \mathcal{C} \to \mathcal{F}$ be a mapping function that projects a solution $s$ to a feature vector $\mathbf{v} = \Phi(s)$. In this work, $\mathbf{v}$ consists of interpretable dimensions, for example, $\mathbf{v} = (\text{Cyclomatic Complexity}, \text{Code Length})$.

\paragraph{MAP-Elites Archive.} We maintain a structured archive (Memory) $\mathcal{A}_{rchive}$, discretized into a grid of cells in $\mathcal{F}$. Each cell, indexed by a feature vector $\mathbf{v}$, stores only the single best solution found so far for that specific behavior:
\begin{equation}
    \mathcal{A}_{rchive}(\mathbf{v}) = \{ s \in \mathcal{C} \mid \Phi(s) \in \text{Cell}(\mathbf{v}) \land R(s) = \max_{s' \in \text{Cell}(\mathbf{v})} R(s') \}
\end{equation}
This mechanism ensures behavioral diversity, preventing the policy from collapsing into a single local optimum.

\subsection{Adaptive Boltzmann Selection}
To select the parent $s_t$ for the next generation from the archive $\mathcal{A}_{rchive}$, we replace static greedy selection with \textbf{Adaptive Boltzmann Selection}.

Let $\{s_1, s_2, \dots, s_N\}$ be the set of solutions currently stored in the archive. The probability $P(s_i)$ of selecting solution $s_i$ as the parent is:
\begin{equation}
    P(s_i) = \frac{\exp(R(s_i) / \tau)}{\sum_{j=1}^{N} \exp(R(s_j) / \tau)}
\end{equation}
where $\tau$ is a temperature parameter dynamically modulated by the population entropy. This allows LoongFlow to shift smoothly between exploration (high $\tau$) and exploitation (low $\tau$).

\section{LoongFlow Overview}
Designing an evolutionary agent capable of solving high-difficulty, open-ended tasks requires overcoming two fundamental systemic contradictions: the tension between search space complexity and sampling efficiency, and the trade-off between population diversity and convergence speed.

\begin{algorithm}[tb]
   \caption{LoongFlow Main Evolutionary Loop}
   \label{alg:LoongFlow_process}
\begin{algorithmic}[1]
   \State {\bfseries Input:} Task Description $T$, Initial Solution $s_0$, Max Iterations $N_{max}$, Islands $K$
   \State {\bfseries Output:} Best Solution $s^*$
   \State \textbf{// Initialization phase}
   \State Initialize Global Memory $\mathcal{M} \leftarrow \{s_0\}$
   \State Initialize $K$ Islands with MAP-Elites Archives $\mathcal{A}_1, \dots, \mathcal{A}_K$
   \For{$iteration = 1$ {\bfseries to} $N_{max}$}
       \For{$k = 1$ {\bfseries to} $K$} \Comment{Parallel Evolution on Islands}
           \State \textbf{// 1. Adaptive Selection (Sec. 4.2.3)}
           \State $H_k \leftarrow \text{CalculateEntropy}(\mathcal{A}_k)$
           \State $\tau \leftarrow \tau_{base} \cdot (1 + \alpha e^{-\beta H_k})$ \Comment{Dynamic Temperature}
           \State $s_{parent} \leftarrow \text{BoltzmannSelect}(\mathcal{A}_k, \tau)$
           \State \textbf{// 2. Lineage-Based Planning (Sec. 4.1.1)}
           \State $chain \leftarrow \text{GetLineage}(s_{parent}.\text{id})$
           \State $context \leftarrow \{ \text{p.plan}, \text{p.summary} \mid p \in chain \}$
           \State $plan \leftarrow \text{Planner}(s_{parent}, context, T)$
           \State \textbf{// 3. Execution \& Evaluation (Sec. 4.1.2)}
           \State $code \leftarrow \text{Executor}(plan, s_{parent})$
           \If{$\text{Verify}(code)$ is False}
               \State \textbf{continue} \Comment{Fast-fail on syntax errors}
           \EndIf
           \State $score, logs \leftarrow \text{Evaluator}(code)$
           \State \textbf{// 4. Reflection \& Storage (Sec. 4.1.3)}
           \State $summary \leftarrow \text{Summarizer}(plan, code, logs)$
           \State $s_{new} \leftarrow \text{Solution}(code, score, summary, parent=s_{parent}.\text{id})$
           \State $\text{UpdateMAPElites}(\mathcal{A}_k, s_{new})$
       \EndFor
       \State \textbf{// 5. Migration Strategy (Sec. 4.2.1)}
       \If{$iteration \mod M == 0$}
           \State \text{MigrateElites}($\mathcal{A}_1, \dots, \mathcal{A}_K$)
       \EndIf
   \EndFor
   \State \textbf{return} $\max_{s \in \cup \mathcal{A}_k} s.score$
\end{algorithmic}
\end{algorithm}

To resolve these, LoongFlow introduces a hierarchical architecture that decouples ``Cognitive Reasoning'' from ``Evolutionary Dynamics''. The framework consists of two coupled subsystems: the \textbf{Agent Loop}, which implements the ``Plan-Execute-Summarize'' (PES) paradigm, and the \textbf{Hybrid Evolutionary Memory}, which governs population management. The overall procedure is outlined in Algorithm~\ref{alg:LoongFlow_process}.

\subsection{The ``Plan-Execute-Summarize'' (PES) Paradigm}
Standard LLM-based evolutionary methods (e.g., genetic programming with LLMs) typically treat the model as a ``black-box mutation operator'', randomly perturbing solutions in hopes of improvement. This approach suffers from extreme sample inefficiency and a lack of directional guidance. To address this, LoongFlow formalizes the evolutionary iteration as a structured cognitive process composed of three specialized stages.

\begin{figure}[htbp]
  \centering
  \includegraphics[width=1.0\linewidth]{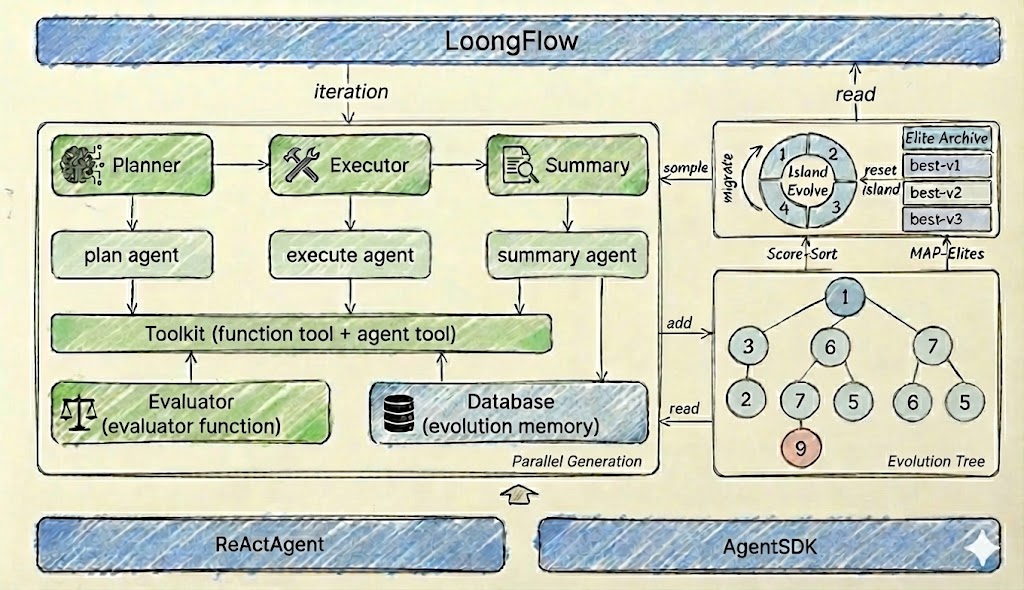}
  \caption{\textbf{Expanded view of the LoongFlow evolutionary process.} The framework iterates through a Planner-Executor-Summarizer loop. The Planner retrieves historical insights to prune the search space; the Executor generates and verifies code; the Summarizer extracts causal knowledge to update the Evolutionary Memory.}
  \label{fig:fig2}
\end{figure}

\subsubsection{Planner: Strategic Search Space Pruning}
In infinite solution spaces, navigating via stochastic mutation often devolves into a ``random walk,'' wasting vast computational resources on invalid or redundant trials. To mitigate this, the \textbf{Planner} functions as a strategic architect employing \textbf{Lineage-Based Context Retrieval}.

Unlike RAG~\cite{lewis2020rag} systems that rely on fuzzy semantic similarity, LoongFlow utilizes the explicit genealogical links inherent in the evolutionary process. As defined in the \texttt{Solution} data structure (see Listing~\ref{lst:solution_struct}), each individual preserves its lineage via \texttt{parent\_id}.

For a given parent solution $s_t$, the Planner traverses the ID chain (retrieving ancestors $s_{t-1}, s_{t-2}...$ and potential descendants). It extracts the historical \texttt{generate\_plan} (Original Intent) and \texttt{summary} (Retrospective Feedback) from this lineage.
\begin{itemize}
    \item \textbf{Intent Tracking}: By reading past plans, the Planner understands the ``research trajectory'' intended by previous generations.
    \item \textbf{Course Correction}: By reading past summaries (which contain specific advice for the next generation), the Planner identifies verified pitfalls to avoid.
\end{itemize}

This structured recall allows the Planner to construct a context-aware blueprint $b$, leveraging the Chain-of-Thought~\cite{wei2022chain} reasoning capabilities of modern LLMs, ensuring that the new plan is a logical continuation and refinement of the parent's strategy, rather than a random jump.

\subsubsection{Executor: Polymorphic Implementation \& Robust Verification}
The translation from a high-level strategic blueprint to an executable solution is inherently non-deterministic and error-prone. The \textbf{Executor} acts as a robust translation engine that converts the Planner's intent ($b$) into verified artifacts ($r$).

\paragraph{Polymorphic Execution Strategies.} The framework decouples the ``What'' (Plan) from the ``How'' (Execution). The Executor supports Pluggable Execution Flows tailored to the problem domain. For algorithmic tasks, it may instantiate as a logic-intensive single-pass coder; for system tasks, it may operate as a multi-stage workflow engine. This design ensures that LoongFlow is not limited to a single class of problems but is adaptable to any domain with a definable action space.

\paragraph{Local Verification Loop (Fast-Fail).} Before submitting to the global Evaluator, the Executor interacts with the Environment Interface to perform ``Pre-Evaluation Checks''. This local feedback loop allows the Executor to self-correct minor errors (such as syntax typos or import errors) immediately, acting as a filter that prevents low-quality candidates from consuming expensive global evaluation resources.

\subsubsection{Summary: Closing the Feedback Loop}
Traditional evolutionary algorithms are ``memory-less'' regarding causality—they know \textit{that} a solution failed, but not \textit{why}. This leads to ``Cyclical Errors,'' where the population repeatedly explores the same invalid dead-ends. The \textbf{Summary} module introduces a retrospective learning mechanism.

After evaluation, the Summarizer performs \textbf{Abductive Reflection}~\cite{shinn2023reflexion}. It compares the Planner's intent ($b$) with the execution result ($r$) to infer causal relationships and generates a structured Insight ($z$). These insights are stored in the Evolutionary Memory. This establishes a \textbf{Long-Term Cognitive Memory}, inspired by the memory architectures in Generative Agents~\cite{park2023generative}, but adapted for evolutionary lineage.. By feeding these insights back to future Planners, LoongFlow achieves Meta-Learning, where the system becomes ``smarter'' about the domain constraints over generations.

\subsection{Hybrid Evolutionary Memory System}
A critical failure mode in evolutionary agents is Premature Convergence. LoongFlow addresses this via a multi-layered memory architecture rooted in a structured data schema.

\paragraph{Solution Data Structure.}
To support the genealogical retrieval described above, LoongFlow maintains a rigorous data schema for every individual in the population. As shown in Listing~\ref{lst:solution_struct}, the \texttt{Solution} class encapsulates not only the code but also the full evolutionary metadata (lineage IDs, plans, summaries, and metrics).

\begin{lstlisting}[language=Python, caption={The Solution Data Structure in Evolutionary Memory}, label={lst:solution_struct}, basicstyle=\ttfamily\footnotesize, frame=single, breaklines=true]
class Solution:
    """Represent a solution in the memory."""

    # Solution identification
    solution: str = ""       # The executable code
    solution_id: str = ""    # Unique UUID

    # Evolution information
    generate_plan: str = ""  # The blueprint that created this solution
    parent_id: Optional[str] = "" # Pointer to ancestor (Lineage Chain)
    island_id: Optional[int] = 0
    iteration: Optional[int] = 0
    timestamp: float = field(default_factory=time.time)
    generation: int = 0
    
    # Sampling Control
    sample_cnt: int = 0      # Visit count for bandit selection
    sample_weight: float = 0.0

    # Performance metrics
    score: Optional[float] = 0.0
    evaluation: Optional[str] = "" # Raw execution logs
    summary: str = ""        # Critical: Advice for the next generation

    # Metadata
    metadata: Dict[str, Any] = field(default_factory=dict)
\end{lstlisting}

This comprehensive schema transforms the memory from a simple ``High Score List'' into a \textbf{Structured Knowledge Graph}, enabling the Planner to query causal relationships (Why did parent X fail?) rather than just outcomes.

\subsubsection{Multi-Island Distributed Topology}
Single-population models are prone to ``dominance,'' where one successful strategy outcompetes all others. LoongFlow employs a \textbf{Multi-Island Model}~\cite{whitley1994cellular} with a Ring Topology. The population is partitioned into $N$ isolated islands. Each island evolves independently, allowing distinct algorithmic ``species'' to cultivate. Migration occurs only when the diversity difference $\Delta D$ between neighbors exceeds a threshold. The top $k\%$ elites are copied to adjacent islands, acting as ``invasive species'' to shake up stagnation. This spatial isolation ensures \textbf{Global Diversity Maintenance}, preventing the system from getting stuck in local optima.

\subsubsection{MAP-Elites with Feature Grids}
Objective-based selection often discards novel but unpolished solutions (``stepping stones''). Within each island, LoongFlow utilizes a \textbf{MAP-Elites}~\cite{mouret2015illuminating} container. Solutions are mapped to a feature grid $\mathcal{A}$ based on behavioral descriptors $\Phi(s)$ (e.g., Code Complexity $\times$ Memory Usage). The system preserves the best individual for each cell in the grid, not just the global best. This guarantees \textbf{Niche Preservation}, providing a diverse ``gene pool'' for the Planner to cross-pollinate.

\subsubsection{Adaptive Boltzmann Selection}
The balance between Exploration and Exploitation is dynamic. LoongFlow implements \textbf{Entropy-Regularized Boltzmann Selection}~\cite{thierens1999scalability}. The selection temperature $\tau$ is dynamically adjusted based on the population entropy $H(\mathcal{P})$:
\begin{equation}
\tau(t) \propto \exp(-\lambda \cdot H(\mathcal{P}_t))
\end{equation}
When the population is diverse (High $H$), $\tau$ lowers to encourage Exploitation (Greedy). When the population converges (Low $H$), $\tau$ rises to force Exploration (Random). This achieves \textbf{Self-Adaptive Control}, automatically transitioning between ``searching for new ideas'' and ``polishing existing ones'' without human intervention.

\section{Experiments}
To empirically validate the Eadfent framework, we conducted a comprehensive evaluation focusing on \textbf{Effectiveness} (Solution Quality) and \textbf{Efficiency} (Convergence Speed).

\subsection{Experimental Setup}
\textbf{Benchmarks}: We instantiated two domain-specific agents—\textit{General Agent} for algorithmic discovery and \textit{Machine Learning Agent} for machine learning engineering—and compared them against state-of-the-art open-source baselines.

\begin{itemize}
    \item \textbf{AlphaEvolve Suite} (General Agent): A suite of challenging open-ended mathematical problems derived from the AlphaEvolve paper~\cite{novikov2025alphaevolve}.
    \item \textbf{MLEBench~\cite{chan2024mle}} (ML Agent): Real-world machine learning competitions requiring end-to-end pipeline optimization, spanning Computer Vision, NLP, and Tabular data.
\end{itemize}

\textbf{Baselines}: We compared General Agent against two primary evolutionary agent frameworks:
\begin{itemize}
    \item \textbf{OpenEvolve}: A standard implementation of the AlphaEvolve algorithm using Island Models.
    \item \textbf{ShinkaEvolve}: A recent framework emphasizing sample efficiency via novelty search.
\end{itemize}

\textbf{Models}: Experiments were conducted using both open-weights models (DeepSeek-r1-0528, etc.) and commercial models (Gemini-3-Pro-Preview, etc.) to ensure the results are framework-dependent rather than model-dependent.

\subsection{Effectiveness}
\subsubsection{Algorithmic Discovery (General Agent)}
We compared the best solutions found by LoongFlow against the baselines and known theoretical bounds. As shown in Table~\ref{tab:alphaevolve}, LoongFlow achieved state-of-the-art (SOTA) results across multiple problems in the benchmark suite.

\begin{table}[htbp]
  \caption{LoongFlow Performance on AlphaEvolve Suite (Grouped by Metric Direction)}
  \label{tab:alphaevolve}
  \centering
  \resizebox{\linewidth}{!}{
  \begin{tabular}{lllll}
    \toprule
    \textbf{Problem}     & \textbf{Metric}     & \textbf{LLM}     & \textbf{AlphaEvolve}      & \textbf{LoongFlow} \\
    \midrule
    \multicolumn{5}{l}{\textit{Higher is Better} ($\uparrow$)} \\
    \midrule
    \textbf{Autocorrelation II} & Bound ($\uparrow$)& DeepSeek-R1 & 0.8962 & \textbf{0.9027} \\
    Circle Packing (Square) & Radius ($\uparrow$) & DeepSeek-R1 & 2.6358 & \textbf{2.6359}      \\
    Circle Packing (Rectangle) & Radius ($\uparrow$) & DeepSeek-R1 & 2.3658321 & \textbf{2.3658322}      \\
    \midrule
    \multicolumn{5}{l}{\textit{Lower is Better} ($\downarrow$)} \\
    \midrule
    Hexagon Packing & Side Length ($\downarrow$) & DeepSeek-R1 & 3.93 & \textbf{3.92}     \\
    Max-to-Min Ratios & Ratio ($\downarrow$)& DeepSeek-R1 & 12.88926 & \textbf{12.88924} \\
    Uncertainty Inequality & Bound ($\downarrow$) & DeepSeek-R1 & 0.352099104422 & \textbf{0.352099104421} \\
    Erd\H{o}s' problem & Bound ($\downarrow$) & DeepSeek-R1 & 0.380924 & \textbf{0.380913} \\
    \bottomrule
  \end{tabular}
  }
\end{table}

Notably, in the \textit{Autocorrelation II} problem, LoongFlow discovered a solution with a score of 0.9027, significantly outperforming the AlphaEvolve baseline (0.8962). This indicates that the Planner's ability to enforce global structural constraints allows LoongFlow to navigate high-dimensional spaces more effectively than random mutation.

\subsubsection{Machine Learning Engineering (ML Agent)}
In the Machine Learning domain, MLAgent demonstrated the ability to construct robust pipelines without human intervention. As shown in Table~\ref{tab:mlebench}, LoongFlow achieved 14 Gold Medals.

\begin{table}[htbp]
  \caption{LoongFlow Performance on MLE Bench}
  \label{tab:mlebench}
  \centering
  \begin{tabular}{lll}
    \toprule
    \textbf{Problem}    & \textbf{LLM}  & \textbf{Result} \\
    \midrule
    Predict-volcanic-eruptions-ingv-oe & Gemini-3.0-flash & \textbf{Gold} \\
    Stanford-covid-vaccine & Gemini-3.0-flash & \textbf{Gold} \\
    The-icml-2013-whale-challenge-right-whale-redux & Gemini-3.0-flash & \textbf{Gold} \\
    Aerial Cactus Identification & Claude-Opus-4.5 & \textbf{Gold} \\
    Nomad2018-predict-transparent-conductors & Claude-Opus-4.5 & \textbf{Gold} \\
    Denoising Dirty Documents & Gemini-3 & \textbf{Gold} \\
    Detecting-insults-in-social-commentary & Gemini-3.0-flash & \textbf{Gold} \\
    Dogs-vs-cats-redux-kernels-edition & Gemini-3.0-flash & \textbf{Gold} \\
    Histopathologic-cancer-detection & Gemini-3.0-flash & \textbf{Gold} \\
    Plant-pathology-2020-fgvc7 & Gemini-3.0-flash & \textbf{Gold} \\
    Tabular-playground-series-dec-2021 & Gemini-3.0-flash & \textbf{Gold} \\
    Google-quest-challenge & Gemini-3.0-flash & \textbf{Gold} \\
    Plant-pathology-2021-fgvc8 & Gemini-3.0-flash & \textbf{Gold} \\
    Us-patent-phrase-to-phrase-matching & Gemini-3.0-flash & \textbf{Gold} \\
    \bottomrule
  \end{tabular}
\end{table}

\subsection{Efficiency and Stability Analysis}
To quantify efficiency, we analyzed the \textit{Circle Packing (Square)} task under strict compute budgets.

\subsubsection{Evolve Efficiency (DeepSeek-R1-0528)}
We set a time limit of 24 hours with a target score $\geq 0.99$. As shown in Table~\ref{tab:efficiency}, LoongFlow demonstrated a $>60\%$ improvement in evolutionary efficiency compared to OpenEvolve.

\begin{itemize}
    \item \textbf{Convergence}: LoongFlow required an average of 258 evaluations to reach the target threshold (0.99), whereas OpenEvolve required 783 evaluations.
    \item \textbf{Success Rate}: Across 3 independent runs, LoongFlow achieved a 100\% success rate in reaching the high-score region ($>0.99$). In contrast, OpenEvolve only succeeded once (33\% rate), and ShinkaEvolve failed to break the 0.99 barrier in all attempts.
\end{itemize}

\begin{table}[htbp]
    \centering
    \caption{Efficiency Comparison (Sorted by Best Score)}
    \label{tab:efficiency}
    \resizebox{\linewidth}{!}{
    \begin{tabular}{lcccccc}
        \toprule
        \textbf{Agent} & \textbf{Round} & \textbf{Best Score} & \textbf{Iter Count} & \textbf{Gen Count} & \textbf{Eval Count} & \textbf{Correctness} \\
        \midrule
        \textbf{LoongFlow} & 2 & \textbf{0.998} & 51 & 484 & 484 & 100\% \\
                       & 1 & 0.996 & 13 & 147 & 147 & 100\% \\
                       & 3 & 0.994 & 28 & 145 & 145 & 100\% \\
        \midrule
        OpenEvolve & 2 & 0.994 & 783 & 783 & 783 & 29.54\% \\
                   & 3 & 0.962 & 1000 & 1000 & 1000 & 48.5\% \\
                   & 1 & 0.950 & 1000 & 1000 & 1000 & 37.9\% \\
        \midrule
        ShinkaEvolve & 3 & 0.952 & 454 & 510 & 454 & 80.84\% \\
                     & 1 & 0.856 & 469 & 568 & 469 & 82.05\% \\
                     & 2 & 0.804 & 300 & 360 & 300 & 77\% \\
        \bottomrule
    \end{tabular}
    }
\end{table}

\subsubsection{High-Difficulty Breakthrough (Gemini-3-Pro)}
Under a constrained budget of 100 iterations, we evaluated the agents' ability to break theoretical barriers. As shown in Table~\ref{tab:breakthrough}, LoongFlow completed the task three times consecutively.

\begin{itemize}
    \item \textbf{LoongFlow}: Successfully broke the theoretical barrier (Score $> 1.0$) in 3 out of 3 runs.
    \item \textbf{Baselines}: Both OpenEvolve and ShinkaEvolve failed to reach a score of 1.0 within the budget. This confirms that LoongFlow's PES paradigm not only finds solutions faster (6 vs 100 calls) but accesses solution subspaces that are unreachable for standard evolutionary methods under limited budgets.
\end{itemize}

\begin{table}[htbp]
    \centering
    \caption{High-Difficulty Breakthrough (Top-100 Iterations)}
    \label{tab:breakthrough}
    \resizebox{\linewidth}{!}{
    \begin{tabular}{lcccccc}
        \toprule
        \textbf{Agent} & \textbf{Round} & \textbf{Best Score} & \textbf{Iter Count} & \textbf{Gen Count} & \textbf{Eval Count} & \textbf{Correctness} \\
        \midrule
        \textbf{LoongFlow} & 1 & \textbf{1.000} & 6 & 20 & 20 & 100\% \\
                       & 2 & \textbf{1.000} & 14 & 72 & 72 & 100\% \\
                       & 3 & \textbf{1.000} & 8 & 26 & 26 & 100\% \\
        \midrule
        OpenEvolve & 1-3 & < 0.998 & 100 & 100 & 100 & 81\% (Avg) \\
        ShinkaEvolve & 1-3 & < 0.999 & 100 & 116 & 100 & 94\% (Avg) \\
        \bottomrule
    \end{tabular}
    }
\end{table}

\subsection{Ablations}
To validate the necessity of the "Plan-Execute-Summarize" (PES) paradigm, we conducted an ablation study using General Agent—the representative instantiation of the LoongFlow framework. We evaluated the contribution of the Planner, Executor, and Summary modules on the Circle Packing task. 

\begin{figure}[htbp]
  \centering
  \begin{minipage}[t]{0.48\textwidth}
    \centering
    \includegraphics[height=4.5cm, keepaspectratio]{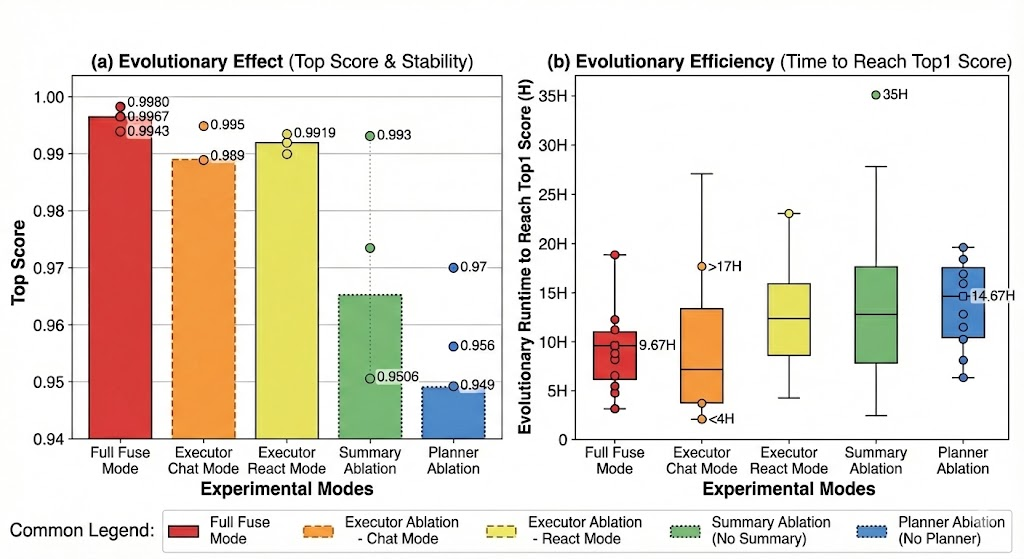}
    \caption{Evolutionary Effect \& Efficiency.}
    \label{fig:ablation_eff}
  \end{minipage}
  \hfill
  \begin{minipage}[t]{0.48\textwidth}
    \centering
    \includegraphics[height=4.5cm, keepaspectratio]{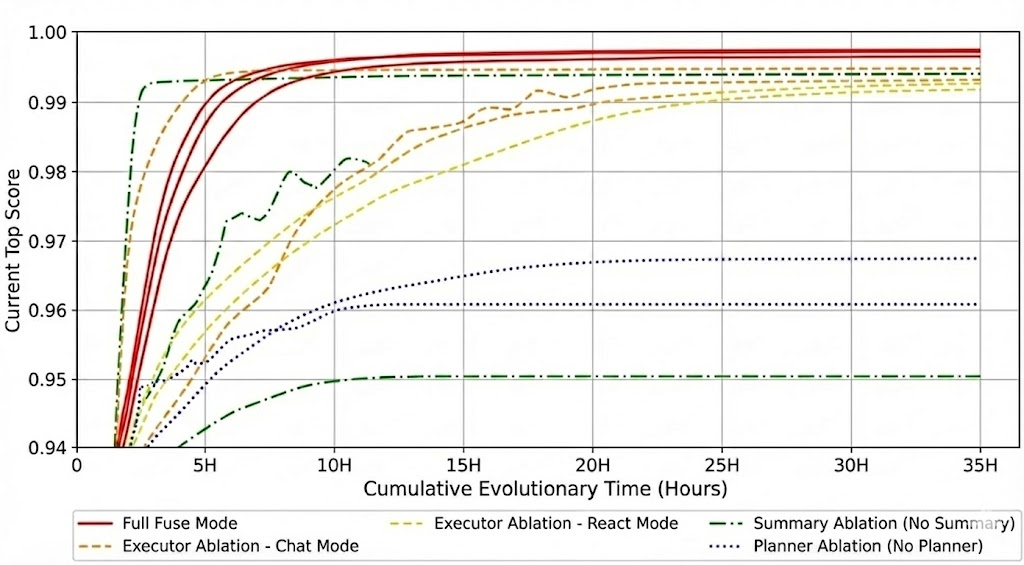}
    \caption{Score Convergence over Time. \textbf{Note}: Faint lines represent individual runs ($N=3$), and bold lines represent the average trajectory.}
    \label{fig:ablation_curve}
  \end{minipage}
\end{figure}

\subsubsection{Planner: The Compass of Evolution}
The \textit{Planner} provides global expert guidance. Removing it forces the agent into a "blind search" mode. As shown in Figure~\ref{fig:ablation_eff}, the Planner-ablated agent stagnated below 0.96. The lack of search pruning increased the average time to reach Top-1 solutions from 9.67 hours to 14.67 hours.

\subsubsection{Executor: Balancing Speed and Depth}
The \textit{Executor} employs an adaptive "Fuse Mode", switching between Chat (single-turn) and ReAct (multi-turn).
\begin{itemize}
    \item \textbf{Chat Mode}: Computationally lightweight but highly unstable.
    \item \textbf{ReAct Mode}: Stable but inefficient.
    \item \textbf{Fuse Mode}: By dynamically allocating compute, Fuse Mode achieved the highest asymptotic score (0.998) with optimal sample efficiency.
\end{itemize}

\subsubsection{Summary: The Evolutionary Feedback}
The \textit{Summary} prevents the loss of historical insights. Without it, the agent suffered from cyclical errors. One trial ran for 35 hours yet failed to break the 0.95 threshold (as shown in Figure~\ref{fig:ablation_curve}, see the "No-Summary" trajectory). The absence of retrospective analysis degraded the Planner's decision-making, confirming that the summary module is essential for breaking performance bottlenecks.

\section{Conclusion}
In this work, we introduced \textbf{LoongFlow}, a cognitive evolutionary framework that fundamentally transcends the "blind watchmaker" limitations of traditional LLM-based optimization. By identifying the critical "cognitive ceiling" in existing methods—specifically their reliance on stochastic mutation and lack of historical reflection—we proposed a paradigm shift from random search to \textbf{Directed Cognitive Evolution}. 

Our core contributions, the \textbf{"Plan-Execute-Summarize" (PES)} paradigm and the \textbf{Hybrid Evolutionary Memory}, effectively bridge the gap between reasoning agents and evolutionary computation. Theoretical analysis and extensive experiments demonstrate that LoongFlow not only preserves the diversity benefits of population-based methods but also injects the strategic depth of reasoning agents, achieving state-of-the-art results with significantly reduced computational overhead. LoongFlow establishes a new standard for sample-efficient, autonomous scientific discovery. 

Future work will focus on extending LoongFlow towards fully autonomous "Meta-Agents" that can self-configure their evolutionary strategies and learning unsupervised diversity metrics for novel domains.

\bibliographystyle{unsrt}  
\bibliography{references}

\end{document}